\documentclass[runningheads]{llncs}
\usepackage{graphicx}
\usepackage{makecell}
\usepackage{cite}
\usepackage{amssymb}
\usepackage{utfsym}
\usepackage{ragged2e}
\usepackage{amsmath}
\usepackage{booktabs}
\usepackage[hidelinks,colorlinks=true,linkcolor=blue,citecolor=blue]{hyperref}
\usepackage{makecell}
\usepackage{colortbl}
\usepackage{xcolor}
\usepackage{float}
\usepackage{authblk}

\begin{document}

\title{GDDS: Pulmonary Bronchioles Segmentation with Group Deep Dense Supervision }
%
\titlerunning{GDDS}
%
\author{Mingyue Zhao\inst{1} \and
Shang Zhao\inst{1}\and
Quan Quan\inst{2}\and
Li Fan\inst{3}\and
Xiaolan Qiu\inst{4}\and
Shiyuan Liu\inst{3}\and
S.Kevin Zhou\inst{1,2}}

%
\institute{Center for Medical Imaging, Robotics, Analytical Computing \& Learning (MIRACLE), School of Biomedical Engineering, University of Science and Technology of China, Suzhou 215123, China \and
Key Lab of Intelligent Information Processing of Chinese Academy of Sciences (CAS), Institute of Computing Technology, CAS, Beijing 100190, China\and
Department of Radiology, Second Affiliated Hospital, Naval Medical University, Shanghai  200003, China \and
Suzhou Key Laboratory of Microwave Imaging, Processing and Application Technology
Suzhou Aerospace Information Research Institute}
%

\maketitle              

\begin{abstract}
Airway segmentation, especially bronchioles segmentation, is an important but challenging task because distal bronchus are sparsely distributed and of a fine scale. Existing neural networks usually exploit sparse topology to learn the connectivity of bronchioles and inefficient shallow features to capture such high-frequency information, leading to the breakage or missed detection of individual thin branches. To address these problems, we contribute a new bronchial segmentation method based on Group Deep Dense Supervision (GDDS) that emphasizes fine-scale bronchioles segmentation in a simple-but-effective manner. First, Deep Dense Supervision (DDS) is proposed by constructing local dense topology skillfully and implementing dense topological learning on a specific shallow feature layer. 
 GDDS further empowers the shallow features with better perception ability to detect bronchioles, even the ones that are not easily discernible to the naked eye. Extensive experiments on the BAS benchmark dataset have shown that our method promotes the network to have a high sensitivity in capturing fine-scale branches and outperforms state-of-the-art methods by a large margin (+12.8\% in BD and +8.8\% in TD) while only introducing a small number of extra parameters.
\keywords{Bronchioles segmentation \and Group deep dense supervision \and Dense topological learning.}
\end{abstract}
\section{Introduction}
Airway segmentation plays an increasingly vital role in the diagnostic and interventional procedures of many lung diseases. Notably, quantitative CT-based bronchi morphological parameters, such as airway lumen diameter, wall thickness and branching patterns, are important disease phenotypes for further understanding of disease progression and therapeutic interventions in chronic obstructive pulmonary disease (COPD)\cite{lancet_COPD_2022} and asthma. Nevertheless, the small size and blurred airway walls of peripheral bronchi make the manual depiction of airway trees time-consuming, error-prone, and overtly subjective. 

Recently, deep learning methods have been widely used in this task~\cite{qin_airwaynet-se_2020,wang_naviairway_2022,Qin_Bronchiole-Sensitive_2020,wu_two-stage_airwaysegmentation_2022,zhang_differentiable_2022,yu_breakBronchiReconstruction_2022}. However, the segmentation of bronchioles remains very challenging for convolutional neural networks (CNNs) due to limited receptive field or shallow features with insufficient attention to detail, showing that individual branches are broken or cannot be detected. On one hand, from the connectivity of detecting bronchioles perspective, some existing methods attempt to introduce topology-preserving strategies~\cite{wang_pointscatter_2022,shit2021cldice,zhang_differentiable_2022,hu2022homotopy_warping,zhang2022progressive} to improve the connectivity of tubular structures like bronchioles,
such as centerline-based guidance~\cite{shit2021cldice,zhang2022progressive,wang_pointscatter_2022}, homotopy warping-based loss~\cite{hu2022homotopy_warping}, etc. We categorize these topology-preserving strategies as sparse topology learning, which usually emphasizes the supervision of individual topology-related key voxel points.
Whereas \emph{sparse topology learning has insufficient power to improve the connectivity of bronchioles segmentation}.
On the other hand, from the sensitivity of detecting bronchioles perspective, some researchers dedicate to improving the sensitivity by optimizing network architectures or introducing new scale-aware loss functions. Qin \emph{et al.} \cite{qin_Tubule-Sensitive_learning_2021} propose feature recalibration and attention distillation to pay more attention to fine structures. Zheng \emph{et al.} \cite{Zheng_Refined_Local_2021} propose a local-imbalance-based weight and BP-based weight enhancement strategy to improve intra-class imbalance. Although these methods have achieved a certain improvement in the detection of bronchioles, \emph{the shallow features of the network are not employed adequately, which really matter to cope with such a scattered and sparse distribution of bronchioles}.

To tackle these two concerns above, we propose a high bronchiole-sensitive segmentation method based on Group Deep Dense Supervision. The contributions of this paper are as follows. 1) To achieve dense topological learning, we propose a simple-but-effective supervision approach, named \textbf{Deep Dense Supervision (DDS)}. We exploit voxel-wise airway annotation to construct local dense topology skillfully and implement dense topological learning in a manner similar to deep supervision. This is the first attempt of dense topology learning on tubular structure segmentation. 2) To strengthen the bronchiole-sensitive feature learning in shallow layers, we combine the DDS with group supervision, forming \textbf{Group Deep Dense Supervision (GDDS)}. GDDS simultaneously improves the sensing power of peripheral bronchial voxels and optimizes the overall connectivity of the airway tree with a few extra parameters. 3) Extensive experiments and standardized metric evaluation comparisons reveal the superiority of our method in bronchioles extraction.
\begin{figure}[htb]
\includegraphics[width=0.95\textwidth]{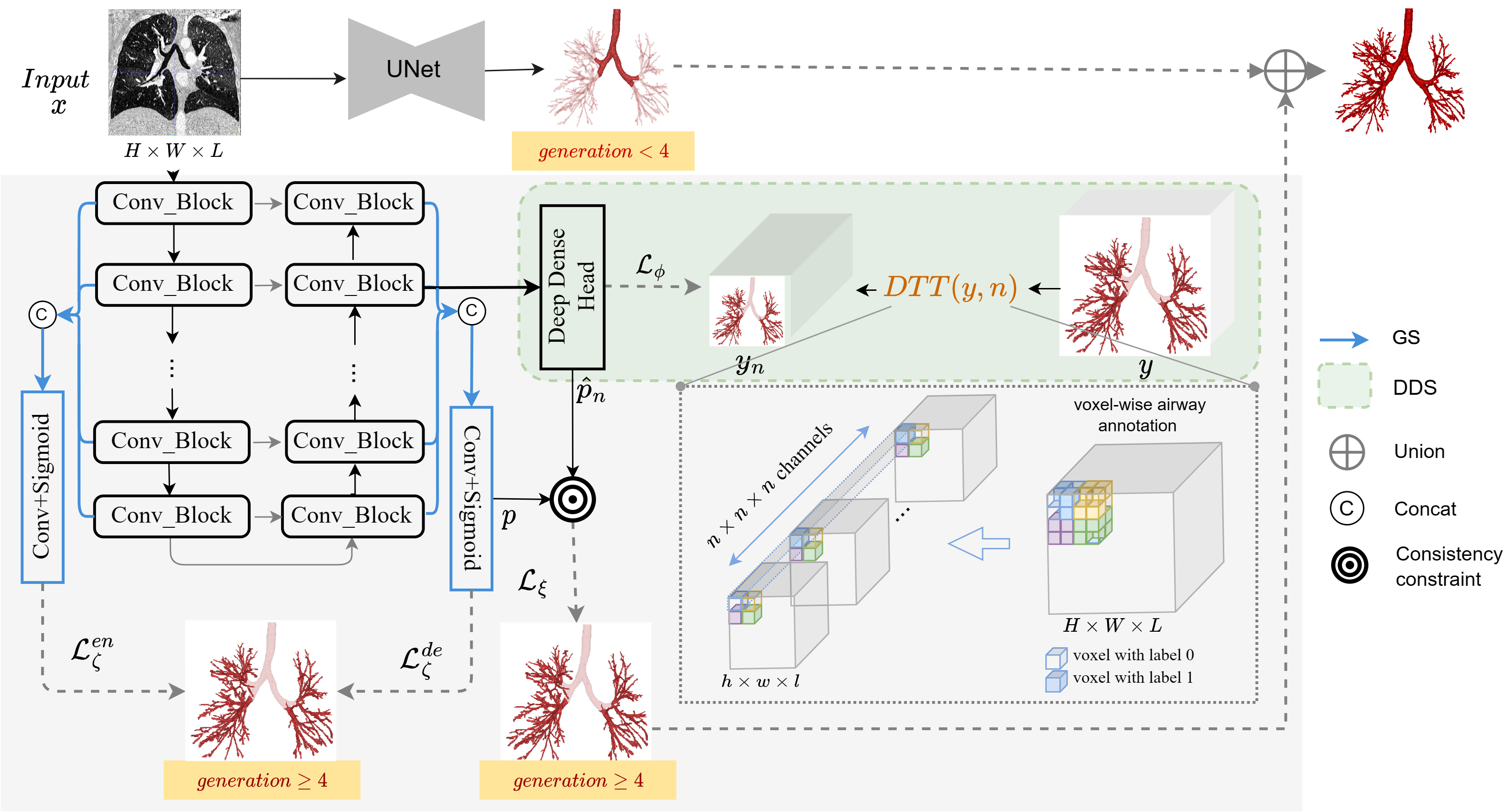}
\caption{Overview of our proposed method for pulmonary airway segmentation.} \label{deep_dense_supervision}
\end{figure}

\section{Methodology} \label{sec:method}
Given the tree structure of the airway, we present a generation-aware training paradigm for overall airway segmentation, as shown in Fig.~\ref{deep_dense_supervision}. 
Referring to \cite{tu_human_2013}, trachea and bronchi lower than $4^{th}$ generation usually measure more than 3mm in diameter and the remaining bronchioles are usually smaller in size, even less than 1mm. In the generation-aware training paradigm, we divide the full label into two parts: i) trachea and bronchi lower than $4^{th}$ generation; ii) the remaining bronchioles.
Subsequently, we perform separate parallel training. The final full-scale airway segmentation result is the combination of the outputs of these two models. Specifically, UNet\cite{3dunet} is exploited to segment the trachea and bronchi. In this paper, we will focus on precise bronchioles segmentation, which will be illustrated in detail. 
%


\subsection{Deep Dense Supervision}
\noindent{\textbf{Dense topological learning.}}
The shallow feature learning of the network is important for the detection of fine structures such as bronchioles. Deep supervision \cite{lee2015deeply} enhances the representation of the low-level features by imposing additional label supervision on different shallow layers. But it may damage hierarchical representation due to the unbiased emphasis on shallow features. This promotes us to ponder: \emph{for the bronchioles segmentation task, which level of shallow layers is more critical? How to design more appropriate label supervision to guide the learning of shallow features?} 
To address the above concerns, we perform dense topological learning on a specific shallow feature layer.


Given an input image $x$ and corresponding airway label $y$ shaped with  $H \times W \times L$. Here we feed the image into the network and obtain the corresponding decoder-side $n \times$down-sampled feature map $f \in \mathbb{R} ^{c\times h \times w \times l}$, where $h$, $w$, $l$ and $c$ indicate the height, width, length and channel dimension, respectively. We have $h = H/n, w = W/n, l =L/n$. Given any point $j$ on the spatial position of the feature map $f$, its feature vector is $\mathbf{\upsilon}_j\in \mathbb{R}^{c\times 1}$. $x^j$ denotes the corresponding  $n \times n\times n$-sized region in the image $x$. Inspired by~\cite{wang_pointscatter_2022}, we argue that the vectors at each spatial position of the down-sampled feature map have a better awareness of local topology. Thus, our goal is to find an effective supervision scheme to enable $\mathbf{\upsilon}_j$ to fully learn the dense topological shape of the bronchioles in area $x^j$.

To achieve this, we first design a dense topological supervision signal on feature map $f$. Specifically, we transform the label $y$ into $y_{n} \in \mathbb{R} ^{n^3 \times h \times w\times l} $ by expanding each cube of size $n \times n \times n$ in $y$ by channel with a stride of $n$, as illustrated in Fig.~\ref{deep_dense_supervision}. Here we define the $n$-fold dense topological transformation on $\kappa$ as $DTT(\kappa,n)$, then we have $y_{n} = DTT(y,n)$.

In addition, a deep dense head following the feature map $f$ is employed to develop the supervision with the local dense topology $y_{n}$. It consists of two $1\times1 \times1 $ convolution blocks and a sigmoid function. The output channel dimension of the deep dense head is $n^3$. Thus, the objective function of the deep dense head is defined as:
\begin{equation}
    \mathcal{L} _{\phi } = FL ( \hat{p}_n ,DTT(y,n)),
\end{equation}
where $\hat{p}_n$ is the predicted probability map of the deep dense head, and $FL$ is the Focal loss~\cite{lin2017focal}.

\begin{figure}[t]
\centering
\includegraphics[width=\textwidth]{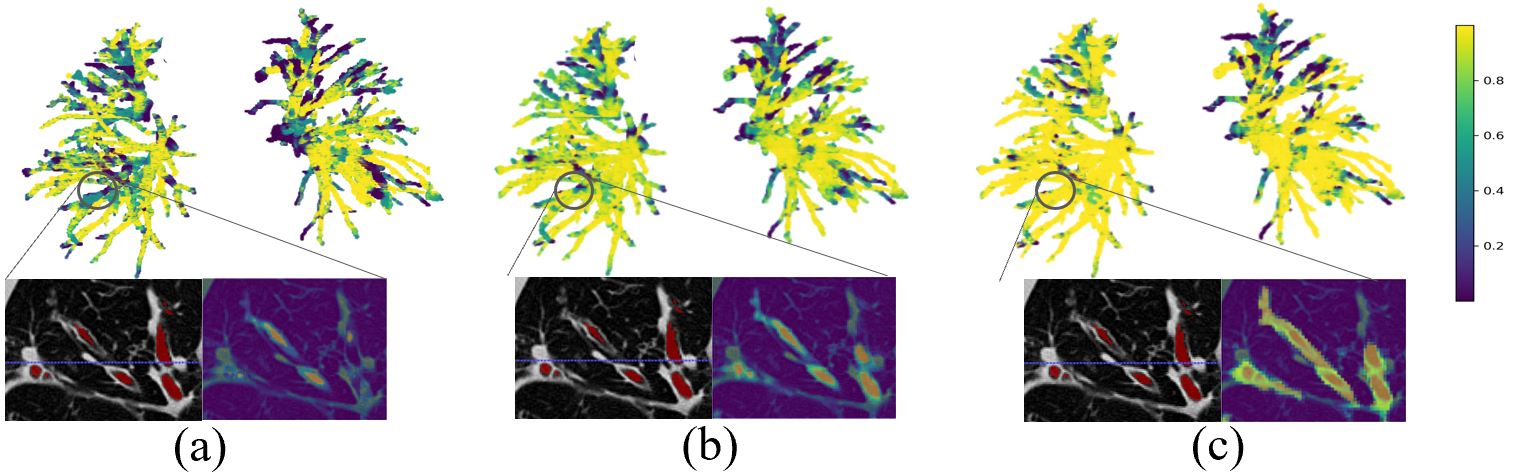}
\caption{Comparsion of probability maps of the segmentation head before DDS (a), the segmentation head after DDS (b) and the deep dense head (c).} \label{activation}
\end{figure}

\noindent{\textbf{Foreground-emphasized consistency constraint.}}
To gain insight into the role of DDS, we perform inverse $DTT$ on the output probability map of the deep dense head, $i.e.$ $\hat{p} = DTT^{-1}(\hat{p}_n,n)$. As shown in Fig.~\ref{activation}, compared with the segmentation head, the deep dense head has a higher response in small airways and low-contrast lumen regions. It enhances the representation of shallow features in fine structure detection and dense topology perception. A comparison of Fig.~\ref{activation}(a) and Fig.~\ref{activation}(b) shows that the addition of DDS boosts the sensitivity of the segmentation layer to these hard regions. In view of this, we further introduce a foreground-emphasized consistency constraint between the output of the deep dense head and the segmentation head, which is formulated as :
\begin{equation}
    \mathcal{L} _{\xi } = FL (y_n*\hat{p}_n*DTT(p,n)+(1-y_n)*((\hat{p}_n+DTT(p,n))/2),y_n),
\end{equation}
where $p$ denotes the predicted probability map of the segmentation head, which is also the output of the whole model. Note that we perform consistency enhancement of the foreground region under the supervision of $y_n$ instead of aligning the outputs of the two heads directly for consistency. In this manner, $\hat{p}_n(i)*DTT(p,n)(i)$ appears as a pseudo-confidence of $i^{th}$ airway voxel under the down-sampled dimension and the loss function $\mathcal{L} _{\xi }$ would penalize heavily on $i$ if it has moderate confidence in any of the prediction map of $\hat{p}_n$ or ${p}_n = DTT(p,n)$.

\subsection{Group Deep Dense Supervision}
On top of DDS, group supervision (GS) is introduced, forming GDDS, to further develop direct supervision enhancement at all levels of the shallow feature layers. 
As illustrated in Fig. \ref{deep_dense_supervision}, feature maps in each convolutional layer in the encoder are upsampled to the input size. Then we form encoder group supervision by concatenating those features. The same goes for the decoding stage. DiceFocal loss is imposed on both stages: 
\begin{equation}
\mathcal{L} _{\zeta } = -(\frac{2 {\textstyle \sum_{i\in x}^{}p(i)y(i)} }{\sum_{i\in x}^{}(p(i)+y(i))+\varepsilon}
+ \frac{1}{\left | x \right |} \sum_{i\in x}^{}(1-p^\prime(i))^2log(p^\prime(i))), 
\end{equation}
where $p^{\prime}(i) = p(i)$ if $y(i) = 1$. Otherwise, $ p^\prime(i) = 1 - p(i)$. $\varepsilon$ is a parameter used to avoid division by zero.
Then, we have the total loss function as follows:
\begin{equation}
\mathcal{L} =\mathcal{L}_{\zeta }^{en}+\mathcal{L}_{\zeta }^{de}+\alpha *\mathcal{L}_{\phi }+ \beta* \mathcal{L}_{\xi },
\end{equation}
where $\mathcal{L}_{\zeta }^{en}$ and $\mathcal{L}_{\zeta }^{de}$ are group supervision loss of encoder and decoder-side respectively. $\alpha$ and $\beta$ are weighting factors of DDS loss and consistency loss. During the inference stage, we only treat the decoder-side GS as the segmentation head, $i.e.$ the output of decoder-side GS is the final segmentation result.


\section{Experiment and Results}
\noindent{\textbf{Datasets.}}
We conduct our experiments on the Binary Airway Segmentation (BAS) Dataset \cite{qin_airwaynet-se_2020}. It consists of 90 CT scans from two public datasets (20 cases from EXACT'09 and 70 cases from LIDC-IDRI). The pixel spacing ranges from 0.5 to 0.82 mm, and the slice thickness ranges from 0.5 to 1.0 mm. In our experiments, 5-fold cross-validation is conducted, meaning that there are 72 cases for training and 18 cases for testing.

\begin{figure}[ht]
\centering
\includegraphics[width=\textwidth]{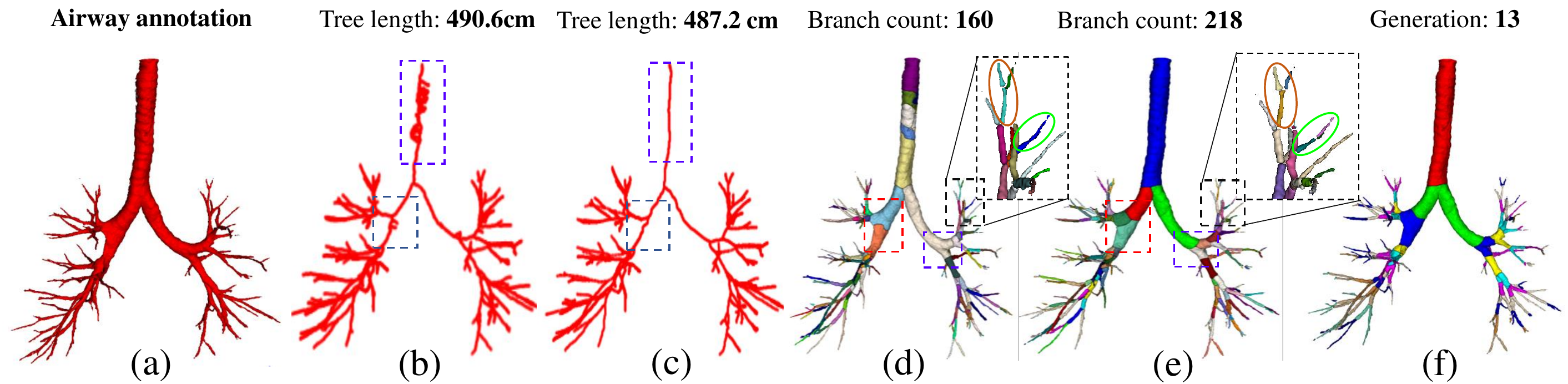}
\caption{Visualisation of airway parsing results. Centerline extraction results of (b)\cite{qin_Tubule-Sensitive_learning_2021} and (c) ours. Branch-level labeling results of (d)\cite{qin_Tubule-Sensitive_learning_2021} and (e) ours. Each color represents a branch. (f) shows the generations of the airway and each color represents a generation.} \label{airway_parsing}
\end{figure}

Accurate centerline extraction and branch-level airway labeling are indispensable for model training and evaluation. 
However, we observe that i) Existing airway parsing methods~\cite{Qin_Bronchiole-Sensitive_2020,qin_airwaynet-se_2020,wang_naviairway_2022} may fail in some cases due to the complex airway morphology. ii) Different parsing methods lead to inconsistent assessment for airway segmentation. Therefore, we implement a more standardized airway parsing. Specifically, the centerlines are first extracted using MIMICs software and then \textit{manually corrected by a panel of well-trained experts}. Branch-level labeling results are obtained according to the parent-child relationship between branch centerlines. Fig. \ref{airway_parsing} shows the comparison between our method and~\cite{qin_Tubule-Sensitive_learning_2021}. Affected by airway morphology and surface smoothness,~\cite{qin_Tubule-Sensitive_learning_2021} produces redundant centerlines in centerline extraction, especially for lower airway branches. This further leads to failures in branch-level parsing, manifested by a branch being split into multiple parts or bifurcation errors between branches. The centerlines and branch labelings obtained by us are more consistent with the actual airway structure and also have great potential for centerline extraction and semantic segmentation tasks. We will \textit{made it public to benefit the community}.

\noindent{\textbf{Implementation Details.}}
The preprocessing includes image value truncation, normalization and DL-based lung extraction~\cite{lungmask_2020}. We apply a sliding window to sample patches from each CT image with the size of $144 \times 144 \times 144$ in low-generation airway segmentation and $80 \times 80 \times 80$ in bronchioles segmentation. On-the-fly data augmentation includes random contrast adjustment, random axis flipping, and random rotation between (-$15^{\circ }, 15^{\circ }$). Our models are trained by Adam optimizer with an initial learning rate of 0.03. The learning rate is divided by 10 in the $20^{th}$, $40^{th}$, and $60^{th}$ epoch. All the experiments are trained until convergence. The testing results are thresholded by 0.5 for binarization. Experimentally, we set $\alpha = \beta = 0.8$ by default. 

\noindent{\textbf{Evaluation Metrics.}}
We adopt volumetric-based and topology-based metrics for evaluation, including Branches Detected (BD), Tree-length Detected (TD), True Positive Rate (TPR), and False Positive Rate (FPR). The definitions can refer to~\cite{qin_Tubule-Sensitive_learning_2021}. Note that: i) Different from  \cite{qin_Tubule-Sensitive_learning_2021}, a branch is considered detected only when 80\% branch voxels are correctly classified; ii) Only the largest component of segmentation results are evaluated on these metrics.

\noindent{\textbf{Quantitative Results.}}
With the backbone of UNet~\cite{3dunet} and WingsNet~\cite{zheng_alleviating_2021}, we compare our method with two classic segmentation networks and four state-of-the-art methods, which are listed in the first column of Table \ref{tab:table8}. All comparison methods except \cite{joint_unet_graph} are implemented according to their official codes and trained from scratch. The values of Juarez \emph{et al}.~\cite{joint_unet_graph} are the reported results in \cite{qin_airwaynet-se_2020}. BD$^{*}$ and TD$^{*}$ measure the branch detected and tree length detected in fine-scale, \emph{i.e.} bronchioles no lower than $4^{th}$ generation in the annotation.

\newcommand{\Frst}[1]{\textcolor{red}{\textbf{#1}}}
\newcommand{\Scnd}[1]{\textcolor{blue}{\textbf{#1}}}
\begin{table*} [h]
    \small
    \caption{Results (\%) of comparison on the testing set of BAS dataset (Mean±Standard deviation). $\uparrow$ indicates that the larger the value, the better the performance. $\downarrow$ is the opposite. \Frst{Red} text and \Scnd{blue} text indicate the best and the second best result, respectively.}
    \label{tab:table8}
    \resizebox{\textwidth}{!}
    {
        \begin{tabular}{l|r|c|c|c|c|c|c}
        \toprule
       
         Method &  \makecell{Params \\ ($\times 10^4$)}  & BD $\uparrow$ &BD${^*} \uparrow$ & TD$\uparrow$ &TD${^*} \uparrow$& TPR $\uparrow$ & FPR $\downarrow$  \\
           
        \midrule
        UNet\cite{3dunet}  & 411.75 & 75.0$^{\pm2.6}$  &   74.1$^{\pm2.8}$&84.5$^{\pm2.5}$ & 82.8$^{\pm2.9}$ & 93.5$^{\pm1.1}$&.023$^{\pm.001}$ \\
        
        Attention UNet\cite{attentionunet}     & 509.91 & 75.6$^{\pm2.3}$  & 74.5$^{\pm2.7}$ &85.1$^{\pm1.9}$&  83.2$^{\pm2.6}$&93.9$^{\pm1.1}$&.023$^{\pm.002}$ \\
        WingsNet\cite{zheng_alleviating_2021}     & 147.31 & 78.4$^{\pm1.9}$  & 77.5$^{\pm2.0}$ &87.6$^{\pm1.4}$&  86.2$^{\pm1.6}$&95.3$^{\pm1.0}$&.042$^{\pm.003}$ \\
        \midrule
        Juarez et al.\cite{joint_unet_graph}  & 5.32      & 77.5$^{\pm20.9}$  & - &66.0$^{\pm20.4}$ & - &77.5$^{\pm15.5}$&\Frst{.009$^{\pm.009}$} \\
        Qin et al. \cite{Qin_Bronchiole-Sensitive_2020}   & 423.12& 76.0$^{\pm{2.6}}$  & 75.1$^{\pm2.7}$ &85.5$^{\pm2.3}$ &   84.0$^{\pm2.5}$& 94.0$^{\pm1.2}$&.023$^{\pm.002}$ \\ 
        
        NaviAirway\cite{wang_naviairway_2022}  & 1076.5 & 68.6$^{\pm2.1}$  & 67.4$^{\pm2.3}$ &	79.7$^{\pm1.8}$ & 77.5$^{\pm2.1}$ &	94.2$^{\pm0.7}$&.061$^{\pm.005}$ \\ 
        
        Zheng et al.\cite{zheng_alleviating_2021} ${^ \dagger}$  & 147.32  & 77.7$^{\pm3.4}$  & 76.8$^{\pm3.6}$ &	87.0$^{\pm2.4}$ &85.5$^{\pm2.8}$ &	92.4$^{\pm1.0}$&\Scnd{.019$^{\pm.002}$} \\
        Proposed(UNet)  & 147.34   & \Scnd{90.4$^{\pm1.9}$}  & \Scnd{90.1$^{\pm2.0}$}&\Scnd{95.7$^{\pm1.4}$} &\Scnd{93.2$^{\pm1.5}$} &	\Scnd{98.3$^{\pm0.7}$}&.128$^{\pm.024}$ \\
        Proposed(WingsNet)  & 147.34   & \Frst{90.5$^{\pm1.5}$}  & \Frst{90.2$^{\pm2.0}$}&\Frst{95.8$^{\pm1.0}$} &\Frst{95.4$^{\pm1.4}$} &	\Frst{98.4$^{\pm0.2}$}&.134$^{\pm.024}$ \\
        \bottomrule
    \end{tabular}
    }
    \begin{flushleft}

        \scriptsize ${^\dagger }$ The result is the average performance after discarding three complete failure cases.
        
    \end{flushleft}
\end{table*}

\begin{table*} [h]
    \small
    \caption{Results (\%) of ablation study on the testing set of BAS dataset (Mean±Standard deviation).}
    \label{tab:ablation}
    \resizebox{\textwidth}{!}
    {
        \begin{tabular}{l|l|c|c|c|c|c|c}
        \toprule
       
         Backbone (WingsNet) & \makecell{Params \\ ($\times 10^4$)}  & BD $\uparrow$  &BD${^*}\uparrow$& TD$\uparrow$ &TD${^*} \uparrow$& TPR $\uparrow$ & FPR $\downarrow$ \\
          
        \midrule

         \makecell[l]{ - \\w/ GT  \\ w/ GT \& DS \\ w/ GT \& DDS($n=8$)\\w/ GT \& DDS($n=4$)\\w/ GT \& DDS($n=2$)\\w/ GT \& GDDS($n=2$)}
        & \makecell[c]{ 147.31 \\ 147.31 \\ 147.37\\153.91 \\147.73 \\147.34 \\147.34} 
        & \makecell[c]{  78.4$^{\pm1.9}$\\ 82.8$^{\pm2.0}$ \\  83.1$^{\pm2.4}$ \\86.2$^{\pm2.1}$ \\ 86.8$^{\pm1.9}$\\ \Scnd{88.6$^{\pm2.0}$} \\\Frst{90.5$^{\pm1.5}$} }
        & \makecell[c]{ 77.5$^{\pm2.0}$ \\82.3$^{\pm2.1}$ \\ 81.7$^{\pm2.9}$ \\ 85.7$^{\pm2.1}$\\86.4$^{\pm2.0}$ \\ \Scnd{88.6$^{\pm3.6}$}\\ \Frst {90.2$^{\pm2.0}$}}  
        & \makecell[c]{ 87.6$^{\pm1.4}$\\ 90.7$^{\pm1.5}$ \\ 91.0$^{\pm2.2}$\\ 93.1$^{\pm1.7}$\\ 93.5$^{\pm1.2}$\\\Scnd{94.6$^{\pm1.5}$}\\ \Frst{95.8$^{\pm1.0}$}}
        & \makecell[c]{ 86.2$^{\pm1.6}$ \\ 89.6$^{\pm1.7}$ \\  88.9$^{\pm2.9}$\\ 92.3$^{\pm2.0}$\\ \Scnd{94.2$^{\pm2.1}$}\\ 94.2$^{\pm2.5}$\\ \Frst{95.4$^{\pm1.4}$}}  
        & \makecell[c]{  95.3$^{\pm1.0}$\\ 96.2$^{\pm0.7}$ \\ 96.3$^{\pm0.9}$\\97.3$^{\pm0.7}$ \\ 97.5$^{\pm0.7}$\\ \Scnd{97.9$^{\pm0.7}$}\\ \Frst{98.4$^{\pm0.2}$}}
        
        & \makecell[c]{  \Frst{.042$^{\pm.003}$}\\ \Scnd{.054$^{\pm.005}$} \\.056$^{\pm.006}$ \\.106$^{\pm033}$ \\.093$^{\pm.014}$ \\ .131$^{\pm.005}$\\.134$^{\pm.024}$}
        \\
        
        \bottomrule
    \end{tabular}
    }
    
\end{table*}

\begin{figure}[htb]
\centering
\includegraphics[width=\textwidth]{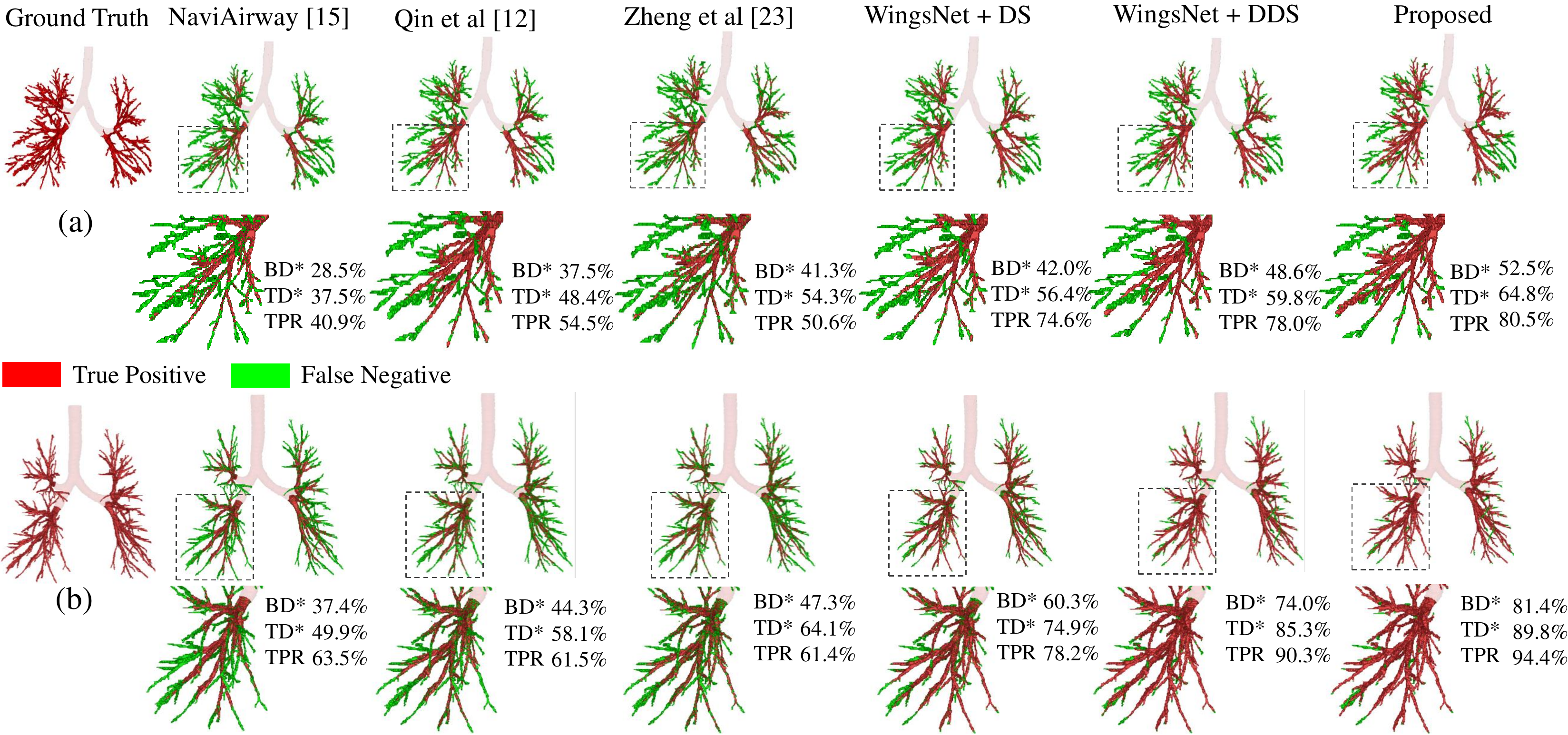}
\caption{Visualisation of segmentation results. (a) is a hard case and (b) is a mild case in the test set of BAS. To emphasize the performance of the proposed method for bronchioles, only TP and FN at fine-scale are shown.} \label{result}
\end{figure}

Results in Table \ref{tab:table8} demonstrate that on the two backbones, GDDS has achieved consistent and substantial improvements. In particular, compared with other methods, GDDS with WingsNet achieves the best performance on BD (90.5\%), BD$^{*}$ (90.2\%), TD (95.8\%), TD$^{*}$(95.4\%), and TPR(98.4\%). This reveals that our proposed method has a better perception of fine structures. We analyze the main causes of false positives generated by algorithms: i) GDDS can detect more real thin branches than doctor annotations; and ii) gradient erosion occurs at some boundaries of the fuzzy lumen and wall of the tube, leading to over-segmentation. 
For reason i), referring to~\cite{wang_naviairway_2022}, we randomly select a $1$-fold test set and calculate the branch ratio (BR), which denotes the number of branches in the model segmentation over the number of branches in the reference segmentation. The average results after review by three experts show that our algorithm can achieve 116.3\% BR. In fact, small airways concentrate more attention from experts in diagnosis. This is because the small airways are frequently involved early in the course of lung disease~\cite{small_airway}.
For example, in COPD, a reduction in the number of small airways and remodeling as important phenotypes allow doctors to intervene early before irreversible pathological changes occur in the lungs~\cite{small_airway_COPD}. Hence, our work has shown some levels of clinical value in assisting doctors to rapidly locates these small airways.

To demonstrate the effectiveness of generation-aware training and GDDS, several comparative studies are carried out in Table \ref{tab:ablation}. Firstly, the significant improvement of results delivered by the generation-aware training confirms that the detection of thin branches requires specialized training strategies. Besides, we compare the effect of different downsample rates $n$ in DDS. $n = 2$ shows the best performance, which outperforms DS with relatively considerable margins (+5.5\% in BD, +6.9\% in BD${^*}$, +3.6\% in TD, and +5.3\% in TD${^*}$). This shows that dense topological supervision strategies based on the specific shallow feature are more effective both in the improvement of connectivity and sensitivity. Furthermore, the addition of group supervision allows GDDS to gain superior performance. It is noteworthy that GDDS significantly improves the sensitivity to small airways with only 0.02\% additional parameters and without any post-processing operations.

\noindent{\textbf{Qualitative Results.}}
Fig.~\ref{result} gives an intuitive performance comparison for the above methods on a hard case and a mild case. Compared with other methods, our algorithms can effectively identify more bronchioles, surpassing others by a large margin.  

\section{Conclusion}
Toward bronchioles segmentation, we propose a simple-but-effective supervision pattern - GDDS. We effectively enhance dense topological perception by performing deep dense supervision. Coupling with group supervision, GDDS further improves the sensing power of bronchioles and optimizes the overall connectivity of the airway tree. Extensive experiments show that our method gains considerable improvement in the detection of bronchioles  with a few extra parameters. Future works include optimizing the boundary segmentation between the airway lumen and the wall, designing a better network for more effective segmentation, and employing airway segmentation for COPD early diagnosis.

%
%

%
%
%
\bibliographystyle{splncs04}
\bibliography{ref}

\begin{thebibliography}{10}
\providecommand{\url}[1]{\texttt{#1}}
\providecommand{\urlprefix}{URL }
\providecommand{\doi}[1]{https://doi.org/#1}

\bibitem{lancet_COPD_2022}
Christenson, S.A., Smith, B.M., Bafadhel, M., Putcha, N.: Chronic obstructive
  pulmonary disease. The Lancet  \textbf{399}(10342),  2227--2242 (Jun 2022)

\bibitem{3dunet}
{\c{C}}i{\c{c}}ek, {\"O}., Abdulkadir, A., Lienkamp, S.S., Brox, T.,
  Ronneberger, O.: 3d u-net: Learning dense volumetric segmentation from sparse
  annotation. In: Ourselin, S., Joskowicz, L., Sabuncu, M.R., Unal, G., Wells,
  W. (eds.) Medical Image Computing and Computer-Assisted Intervention --
  MICCAI 2016. pp. 424--432. Springer International Publishing, Cham (2016)

\bibitem{joint_unet_graph}
Garcia-Uceda~Juarez, A., Selvan, R., Saghir, Z., de~Bruijne, M.: A {Joint} {3D}
  {UNet}-{Graph} {Neural} {Network}-{Based} {Method} for {Airway}
  {Segmentation} from {Chest} {CTs}. In: Suk, H.I., Liu, M., Yan, P., Lian, C.
  (eds.) Machine {Learning} in {Medical} {Imaging}, vol. 11861, pp. 583--591.
  Springer International Publishing, Cham (2019)

\bibitem{lungmask_2020}
Hofmanninger, J., Prayer, F., Pan, J., Röhrich, S., Prosch, H., Langs, G.:
  Automatic lung segmentation in routine imaging is primarily a data diversity
  problem, not a methodology problem. Eur Radiol Exp  \textbf{4}(1), ~50 (Dec
  2020)

\bibitem{hu2022homotopy_warping}
Hu, X.: Structure-aware image segmentation with homotopy warping. In:
  Thirty-sixth Conference on Neural Information Processing Systems (NeurIPS)
  (2022)

\bibitem{lee2015deeply}
Lee, C.Y., Xie, S., Gallagher, P., Zhang, Z., Tu, Z.: Deeply-supervised nets.
  In: Artificial intelligence and statistics. pp. 562--570. PMLR (2015)

\bibitem{lin2017focal}
Lin, T.Y., Goyal, P., Girshick, R., He, K., Doll{\'a}r, P.: Focal loss for
  dense object detection. In: Proceedings of the IEEE international conference
  on computer vision. pp. 2980--2988 (2017)

\bibitem{attentionunet}
Oktay, O., Schlemper, J., Folgoc, L.L., Lee, M., Heinrich, M., Misawa, K.,
  Mori, K., McDonagh, S., Hammerla, N.Y., Kainz, B., et~al.: Attention u-net:
  Learning where to look for the pancreas. arXiv preprint arXiv:1804.03999
  (2018)

\bibitem{small_airway_COPD}
Polverino, F., Soriano, J.B.: Small airways and early origins of copd:
  pathobiological and epidemiological considerations (2020)

\bibitem{qin_airwaynet-se_2020}
Qin, Y., Gu, Y., Zheng, H., Chen, M., Yang, J., Zhu, Y.M.: {AirwayNet}-{SE}:
  {A} {Simple}-{Yet}-{Effective} {Approach} to {Improve} {Airway}
  {Segmentation} {Using} {Context} {Scale} {Fusion}. In: 2020 {IEEE} 17th
  {International} {Symposium} on {Biomedical} {Imaging} ({ISBI}). pp. 809--813.
  IEEE, Iowa City, IA, USA (Apr 2020)

\bibitem{qin_Tubule-Sensitive_learning_2021}
Qin, Y., Zheng, H., Gu, Y., Huang, X., Yang, J., Wang, L., Yao, F., Zhu, Y.M.,
  Yang, G.Z.: Learning {Tubule}-{Sensitive} {CNNs} for {Pulmonary} {Airway} and
  {Artery}-{Vein} {Segmentation} in {CT}. IEEE Trans. Med. Imaging
  \textbf{40}(6),  1603--1617 (Jun 2021)

\bibitem{Qin_Bronchiole-Sensitive_2020}
Qin, Y., Zheng, H., Gu, Y., Huang, X., Yang, J., Wang, L., Zhu, Y.M.: Learning
  {Bronchiole}-{Sensitive} {Airway} {Segmentation} {CNNs} by {Feature}
  {Recalibration} and {Attention} {Distillation}. In: Martel, A.L.,
  Abolmaesumi, P., Stoyanov, D., Mateus, D., Zuluaga, M.A., Zhou, S.K.,
  Racoceanu, D., Joskowicz, L. (eds.) Medical {Image} {Computing} and
  {Computer} {Assisted} {Intervention} – {MICCAI} 2020, vol. 12261, pp.
  221--231. Springer International Publishing, Cham (2020)

\bibitem{shit2021cldice}
Shit, S., Paetzold, J.C., Sekuboyina, A., Ezhov, I., Unger, A., Zhylka, A.,
  Pluim, J.P., Bauer, U., Menze, B.H.: cldice-a novel topology-preserving loss
  function for tubular structure segmentation. In: Proceedings of the IEEE/CVF
  Conference on Computer Vision and Pattern Recognition. pp. 16560--16569
  (2021)

\bibitem{tu_human_2013}
Tu, J., Inthavong, K., Ahmadi, G.: The {Human} {Respiratory} {System}. In:
  Computational {Fluid} and {Particle} {Dynamics} in the {Human} {Respiratory}
  {System}, pp. 19--44. Springer Netherlands, Dordrecht (2013)

\bibitem{wang_naviairway_2022}
Wang, A., Tam, T.C.C., Poon, H.M., Yu, K.C., Lee, W.N.: {NaviAirway}: a
  {Bronchiole}-sensitive {Deep} {Learning}-based {Airway} {Segmentation}
  {Pipeline} (Jun 2022)

\bibitem{wang_pointscatter_2022}
Wang, D., Zhang, Z., Zhao, Z., Liu, Y., Chen, Y., Wang, L.: {PointScatter}:
  {Point} {Set} {Representation} for {Tubular} {Structure} {Extraction}. In:
  Avidan, S., Brostow, G., Cissé, M., Farinella, G.M., Hassner, T. (eds.)
  Computer {Vision} – {ECCV} 2022. pp. 366--383. Springer Nature Switzerland,
  Cham (2022)

\bibitem{wu_two-stage_airwaysegmentation_2022}
Wu, Y., Zhao, S., Qi, S., Feng, J., Pang, H., Chang, R., Bai, L., Li, M., Xia,
  S., Qian, W., Ren, H.: Two-stage {Contextual} {Transformer}-based
  {Convolutional} {Neural} {Network} for {Airway} {Extraction} from {CT}
  {Images} (Dec 2022)

\bibitem{small_airway}
Xiao, D., Chen, Z., Wu, S., Huang, K., Xu, J., Yang, L., Xu, Y., Zhang, X.,
  Bai, C., Kang, J., et~al.: Prevalence and risk factors of small airway
  dysfunction, and association with smoking, in china: findings from a national
  cross-sectional study. The Lancet Respiratory Medicine  \textbf{8}(11),
  1081--1093 (2020)

\bibitem{yu_breakBronchiReconstruction_2022}
Yu, W., Zheng, H., Zhang, M., Zhang, H., Sun, J., Yang, J.: Break: {Bronchi}
  {Reconstruction} by {Geodesic} {Transformation} and {Skeleton} {Embedding}.
  In: 2022 {IEEE} 19th {International} {Symposium} on {Biomedical} {Imaging}
  ({ISBI}). pp.~1--5. IEEE, Kolkata, India (Mar 2022)

\bibitem{zhang_differentiable_2022}
Zhang, M., Yang, G.Z., Gu, Y.: Differentiable {Topology}-{Preserved} {Distance}
  {Transform} for {Pulmonary} {Airway} {Segmentation} (Oct 2022)

\bibitem{zhang2022progressive}
Zhang, X., Zhang, J., Ma, L., Xue, P., Hu, Y., Wu, D., Zhan, Y., Feng, J.,
  Shen, D.: Progressive deep segmentation of coronary artery via hierarchical
  topology learning. In: Medical Image Computing and Computer Assisted
  Intervention--MICCAI 2022: 25th International Conference, Singapore,
  September 18--22, 2022, Proceedings, Part V. pp. 391--400. Springer (2022)

\bibitem{Zheng_Refined_Local_2021}
Zheng, H., Qin, Y., Gu, Y., Xie, F., Sun, J., Yang, J., Yang, G.Z.: Refined
  {Local}-imbalance-based {Weight} for {Airway} {Segmentation} in {CT}. In:
  de~Bruijne, M., Cattin, P.C., Cotin, S., Padoy, N., Speidel, S., Zheng, Y.,
  Essert, C. (eds.) Medical {Image} {Computing} and {Computer} {Assisted}
  {Intervention} – {MICCAI} 2021, vol. 12901, pp. 410--419. Springer
  International Publishing, Cham (2021), series Title: Lecture Notes in
  Computer Science

\bibitem{zheng_alleviating_2021}
Zheng, H., Qin, Y., Gu, Y., Xie, F., Yang, J., Sun, J., Yang, G.Z.: Alleviating
  {Class}-{Wise} {Gradient} {Imbalance} for {Pulmonary} {Airway}
  {Segmentation}. IEEE Trans. Med. Imaging  \textbf{40}(9),  2452--2462 (Sep
  2021)

\end{thebibliography}
%




\end{document}